\documentclass[11pt,a4paper]{article}
\usepackage[hyperref]{acl2021}
\usepackage{times}
\usepackage{latexsym}
\usepackage{soul}
\usepackage{amstext}

\usepackage{natbib}
\usepackage{wrapfig}
\usepackage[disable=true]{todonotes}
\usepackage[font={small, it}]{caption}

\usepackage{microtype}
\usepackage{lipsum}
\usepackage{multicol}
\usepackage{makecell}

\aclfinalcopy

\title{Measuring and Improving BERT's Mathematical Abilities by Predicting the Order of Reasoning}

\author{Piotr Piekos \\
   University of Warsaw\\
   \And
  Henryk Michalewski \\ 
  University of Warsaw, Google\\
  
  \And
  Mateusz Malinowski \\
  DeepMind
  }

\date{}

\begin{document}
\maketitle
\begin{abstract}
Imagine you are in a supermarket. You have two bananas in your basket and want to buy four apples. How many fruits do you have in total? This seemingly straightforward question can be challenging for data-driven language models, even if trained at scale. However, we would expect such generic language models to possess some mathematical abilities in addition to typical linguistic competence. 
Towards this goal, we investigate if a commonly used language model, BERT, possesses such mathematical abilities and, if so, to what degree. For that, we fine-tune BERT on a popular dataset for word math problems, AQuA-RAT, and conduct several tests to understand learned representations better.\newline
Since we teach models trained on natural language to do formal mathematics, we hypothesize that such models would benefit from training on semi-formal steps that explain how math results are derived. To better accommodate such training, we also propose new pretext tasks for learning mathematical rules. We call them (Neighbor) Reasoning Order Prediction (ROP or NROP). With this new model, we achieve significantly better outcomes than data-driven baselines and even on-par with more tailored models. We also show how to reduce positional bias in such  models.
\end{abstract}

\section{Introduction}
Automatically solving math word problems has a long history dating back to the middle sixties \cite{bobrow}.  Early approaches were rule-based matching systems
that 
solve the problem symbolically.  Even though there are some impressive symbolic systems that operate in a relatively narrow domain, the inability to successfully scale them up is sometimes presented as a critique of the good-old-fashioned AI, or GOFAI~\cite{dreyfus1992computers}. One issue is to create a formalism that covers all the aspects needed to solve these problems. On the other hand, deep learning~\cite{lecun2015deep} aims to develop 
artificial general intelligence that scales better to various problems.

However, despite many successes in computer vision and natural language processing~\cite{bert,he2016identity,krizhevsky2012imagenet,lan2019albert,mikolov2013distributed},  data-driven methods evade our dream of building a system with basic, every-day, mathematical skills. 
As large-scale natural language models become more common~\cite{bert,brown2020language}, we would expect them to also reason mathematically.

Since natural language understanding also involves symbolic manipulation~\cite{liang2016learning}, we treat mathematical reasoning as a language understanding and revisit the data-driven paradigm. For that, we rely on a recent language model, BERT~\cite{devlin-etal-2019-bert}, and challenge it with math word problems~\cite{aqua}.
Even though such language models have initially shown promising results, more recent investigation shows they may rely on various biases in their predictions~\cite{hendricks2018women,brown2020language,bhardwaj2020investigating,kurita2019measuring}. Here, we also follow that line of investigation and show these models can answer correctly without an understanding of the rationale behind it.

Furthermore,
as directly predicting answers to
math problems often requires multiple steps of reasoning, we show that we can improve BERT's generalization by exposing it to rationales~\cite{aqua,hendricks2016generating,lei2016rationalizing}. These are, however, only used during training similarly to a teacher that shows a student a justification for each answer. But then, the student is evaluated only on the ability to answer these questions during the college exam correctly with no access to rationales.
Finally, to learn a better representation from rationales and to improve the generalization even further, we introduce novel pretext tasks and corresponding losses, which we name (Neighbor) Reasoning Order Prediction (ROP or NROP). We also show that permutation invariant losses can lead to less biased representations.
With that, we  outperform other data-driven baselines, and are even on-par with methods that are more tailored to math-world problems and the AQuA-RAT dataset.

\begin{figure}[t]
\begin{center}
\includegraphics[width=0.4\textwidth]{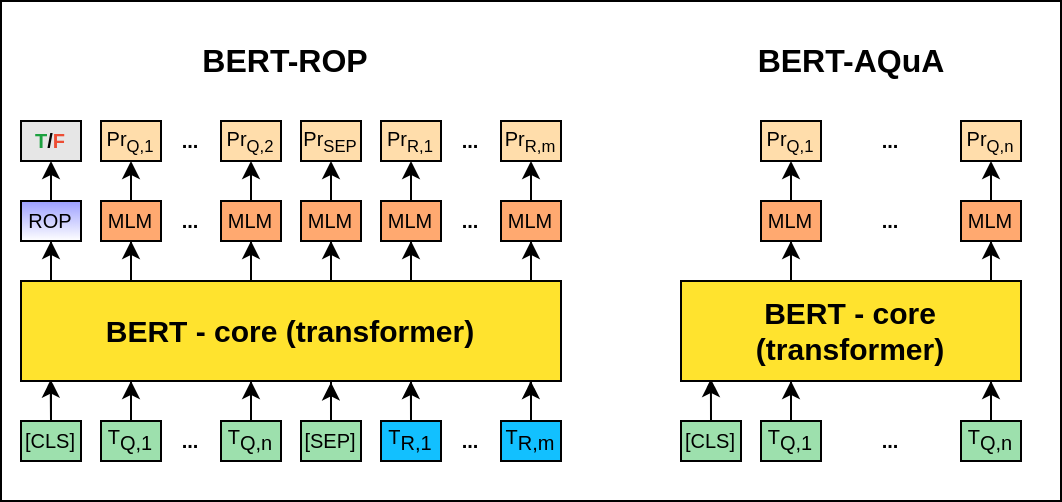}
\caption{
BERT (right) and our novel extension (left).
We use shared architecture but we separate question tokens (green blocks) from rationales (blue blocks) using different segment and positional embeddings.
We show all three losses. 
MLM predicts masked tokens (depicted here as $Pr_{Q,k}$). We use ROP or NROP to predict if the ordering of rationale steps is correct. For question-answering,  we fine-tune the whole model with a classification layer using softmax. We use the embedding that corresponds to the [CLS] token as the input representation.
}
\label{fig:architectures}
\end{center}
\end{figure}

\section{Methods}

We use the following methods, each initialized with BERT-base pre-trained on Wikipedia and Books Corpus~\cite{bert, zhu2015aligning}.
Note that, in fine-tuning they all have the same number of parameters.
\newline
\noindent 1) {\bf BERT-base}. We fine-tune BERT to predict the correct answer and show 
its transfer to math word problems.
\newline
\noindent
2) {\bf BERT-AQuA}. We use the MLM loss on the AQuA-RAT questions before training to predict correct answer.
\newline
\noindent
3) {\bf BERT-AQuA-RAT}. We use the MLM loss on the AQuA-RAT questions and rationales and show if we can inject knowledge from rationales into BERT. 
\newline
\noindent
4) {\bf BERT-(N)ROP}. We use the MLM loss and the novel (N)ROP loss for coherence prediction (defined later) and show if we can improve the results by focusing the model on rationales. 

Later in this paper, we propose permutation invariant losses that additionally reduce positional biases of the BERT-base model, and can work with all the pretext tasks described above.

\subsection{Architectures, pretext tasks and losses}
We base our architecture on BERT~\cite{devlin-etal-2019-bert} that has 12 transformer blocks~\cite{vaswani2017attention}. As the core, we use the standard configuration described in~\cite{devlin-etal-2019-bert}.
We use three self-supervised losses. One is the standard Masked Language Modelling (MLM) but extended to work on rationales. Other two are our new losses, (Neighbour) Reasoning Order Prediction (ROP or NROP).
Figure~\ref{fig:architectures} shows two variants of our models. Note that, during fine-tuning, rationales and all the self-supervised losses are discarded. 

\noindent\textbf{MLM} is the Masked Language Modelling \cite{devlin-etal-2019-bert}. 
We randomly mask $15\%$ of the input tokens by a special token [MASK]. The objective of this loss is to predict the masked token using its context casted as a classification problem over the tokenizer vocabulary. 
Loss is calculated only on masked tokens. We extend this loss to rationales. First, we randomly choose whether we mask a question or rationale. Next, we follow the procedure above applied to either a question or rationale. However, to encourage binding between questions and rationales, we use the whole context for the predictions. 
Interestingly, there are parallels between masking numbers and solving mathematical equations, where it can be seen as solving the equation with unknown.
For example, $2 + [\text{MASK}] = 4$ becomes 
$2 + x = 4$. As a consequence, models during training organically deal with mathematical calculations without defining a specific loss for mathematics 
allowing soft transitions between natural and more formal languages.
\begin{figure}[t]
\centering
    \includegraphics[width=0.85\linewidth]{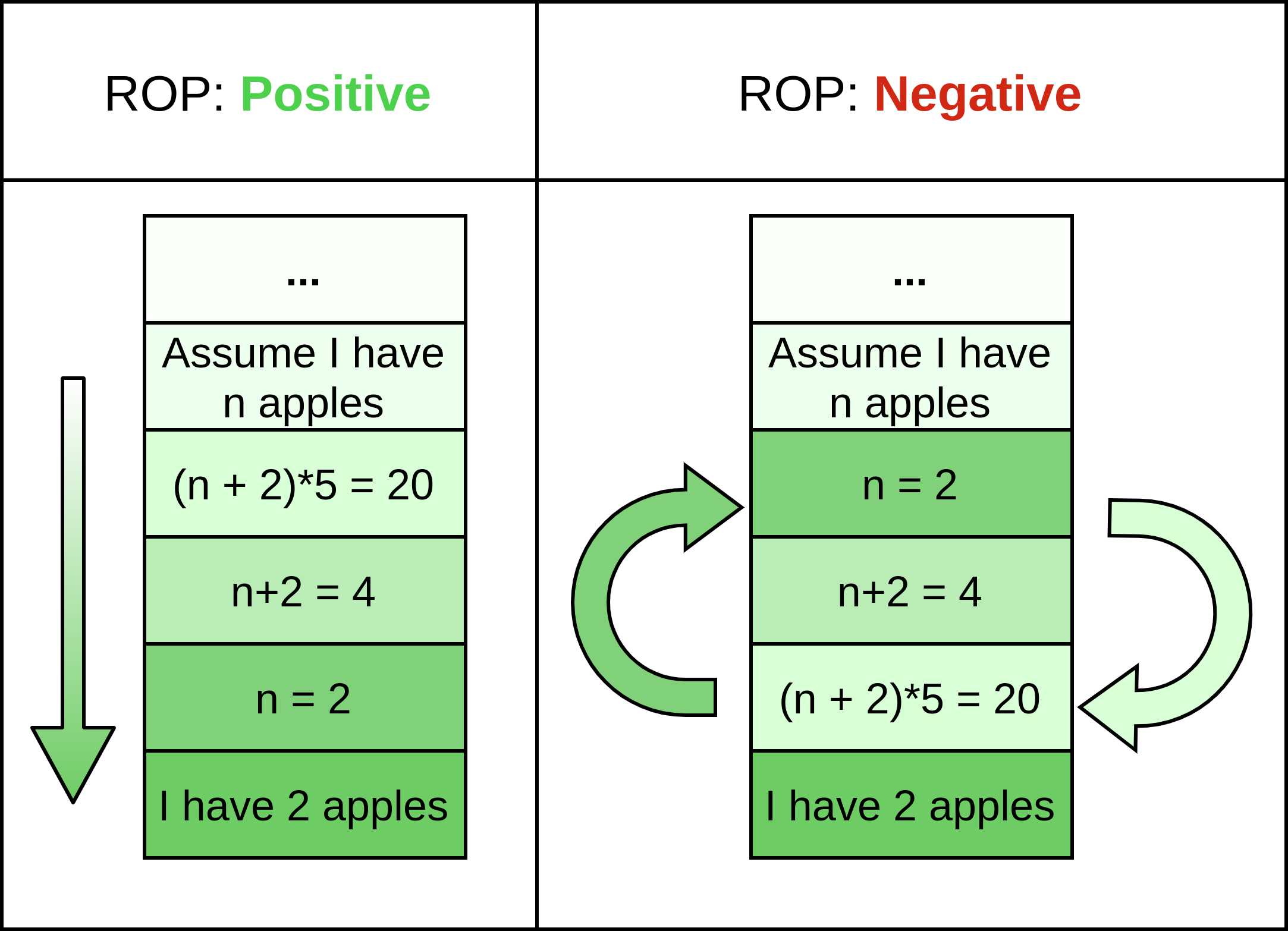} 
    \caption{ROP or NROP with positive (left) and negative (right) labels. We randomly swap two rationales and classify if that change has happened.
    }
    \label{fig:rop}
\end{figure}

\noindent\textbf{ROP} is our novel coherence loss. Since rationales are sequences of consecutive reasoning steps, the order of the execution is critical as shown in Figure~\ref{fig:rop}. Following this intuition, we introduce Reasoning Order Prediction (ROP) that predicts whether the order of the rationale steps is preserved. Hence it encourages the network to pay more attention to rationales. 
The loss is similar to Sentence Order Prediction (SOP)~\cite{lan2019albert}, but ours is focused on learning reasoning steps.
\noindent\textbf{NROP} is an extension of ROP where only consecutive rationale steps are swapped making the prediction (swap or no swap) task more challenging
and, hence, it can arguably lead to a better representation as understanding the correct ordering is more nuanced. 
Indeed, we observe that our models trained with NROP correctly predict if swap has occurred in about $75\%$ cases, while with ROP in about $78\%$ cases (both on the validation set). This indeed, confirms our hypothesis that NROP task is more challenging than ROP.

\section{Results}
\noindent\textbf{Dataset.} We use AQuA-RAT~\cite{aqua}. It has about 100k crowd-sourced math questions with five candidate answers (one is correct). Each question has a rationale -- a step-by-step explanation of how the answer is computed -- that is only available during training. At test time answer predictions are based  on questions. The train set has roughly 100k question-answer-rationale triples, while dev and test about 250 question-answer pairs each.
\newline\noindent
\textbf{Main results.} Table~\ref{table:ta1} shows our main results. 
We see that our method is the state-of-the-art among the models with minimal inductive biases
and is very competitive to the other two models that are more specific to handle word math problems (e.g., requires programs).
Moreover, even though BERT is already a stronger model than LSTM, it is better to use its MLM pretext task and loss on the AQuA-RAT questions (BERT-AQuA) or even better on questions and rationales (BERT-AQuA-RAT). However, models with our novel coherence prediction losses can better learn from rationales (BERT-ROP and BERT-NROP).

\begin{figure}[bt]
\centering
    \includegraphics[width=1\linewidth]{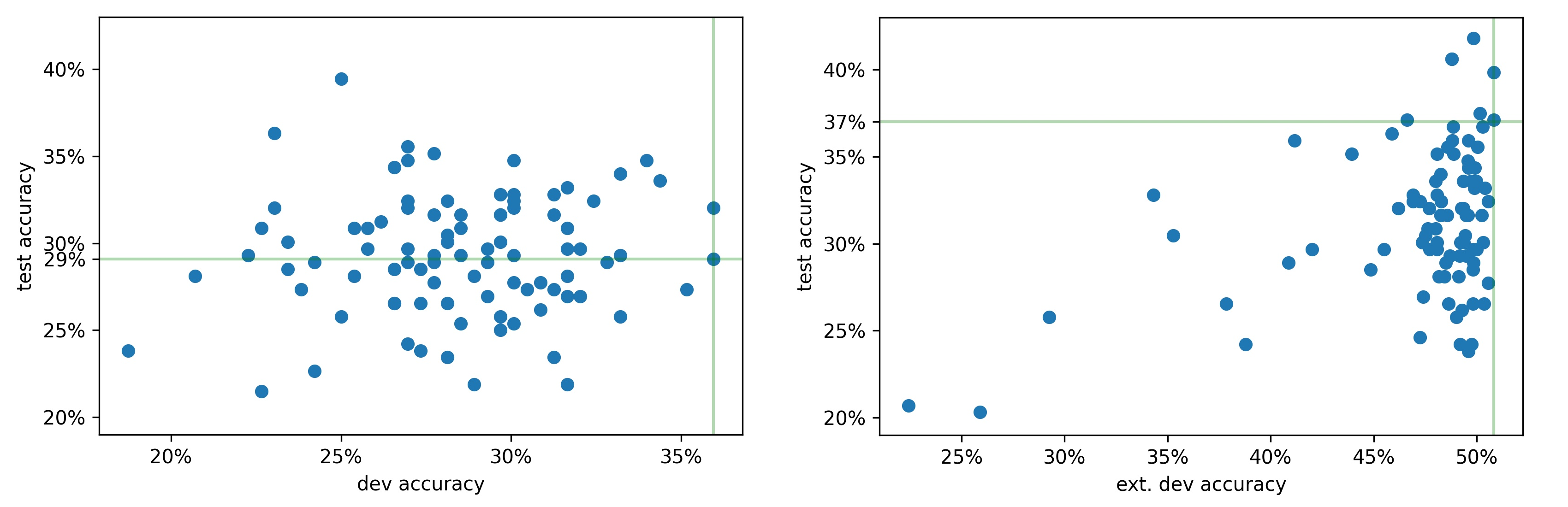} 
    \caption{Accuracies for dev and test sets. Green lines show the iteration that maximizes validation accuracy. The image also shows the sensitivity of relationship between test and the original (left) or our extended (right) validation set.
    }
    \label{fig:val_scatter}
\end{figure}

Moreover, we observe a highly sensitive relationship between dev and test sets (Figure~\ref{fig:val_scatter}, left), where small changes in the accuracies in the former set can lead to more dramatic changes at test time. Indeed, the correlation of results between both sets is only $0.082$. As the validation set is quite small,
we propose an extended dev consisting of 5000 randomly chosen samples from the training set extended by the whole dev set. Although not ideal, and the sensitive relationship is still present (Figure~\ref{fig:val_scatter}, right), we have increased the correlation to $0.401$. With such a new validation set, we report $37\%$ test accuracy but we can also see that $40\%$ is within the reach (Figure~\ref{fig:val_scatter}, right).\todo{add to table}

\begin{table}[tb]
\centering
\resizebox{0.8\linewidth}{!}{
\begin{tabular}{ |p{3.8cm}||p{2cm}|  }
 \hline
\footnotesize{ Model} & \footnotesize{Accuracy}\\
 \hline
 
 \footnotesize{Random chance} & \footnotesize{\ \ \ 20.0\%} \\
  \footnotesize{LSTM~\cite{aqua}}   & \footnotesize{\ \ \ 20.8\%}  \\
  \footnotesize{BERT-base (ours)} &   \footnotesize{\ \ \ $28.3(\pm2.0)\%$}\\
  \footnotesize{BERT-AQUA (ours)} & \footnotesize{\ \ \ $29.1 (\pm 1.7)\%$}\\
  \footnotesize{BERT-AQuA-RAT (ours)} & \footnotesize{ \ \ \  $32.3 (\pm 1.8)\%$ }\\
  \footnotesize{BERT-ROP (ours)} & \footnotesize{ \ \ \  $35.4 (\pm 1.0)\%$}\\
 \footnotesize{\bf BERT-NROP (ours) } & \footnotesize{\ \ \ {\bf $37.0 (\pm 1.1)\%$}} \\
  \hline
\footnotesize{AQuA-RAT~\cite{aqua}} & \footnotesize{\ \ \ 36.4\%} \\
\small{\bf MathQA}~\cite{amini-etal-2019-mathqa} & \small{\bf \ \ \ 37.9\%} \\
\hline
\end{tabular}
}
\caption{Comparison of data-driven (first six rows) with two hybrid approaches that use stronger and hence more specific inductive biases (last two rows). Standard deviation estimates (over random initializations) is given in parentheses, where we see our losses can reduce the variability slightly.
}
\label{table:ta1}
\end{table}

\begin{figure}[tb]
  \centering
   \includegraphics[width=0.8\linewidth]{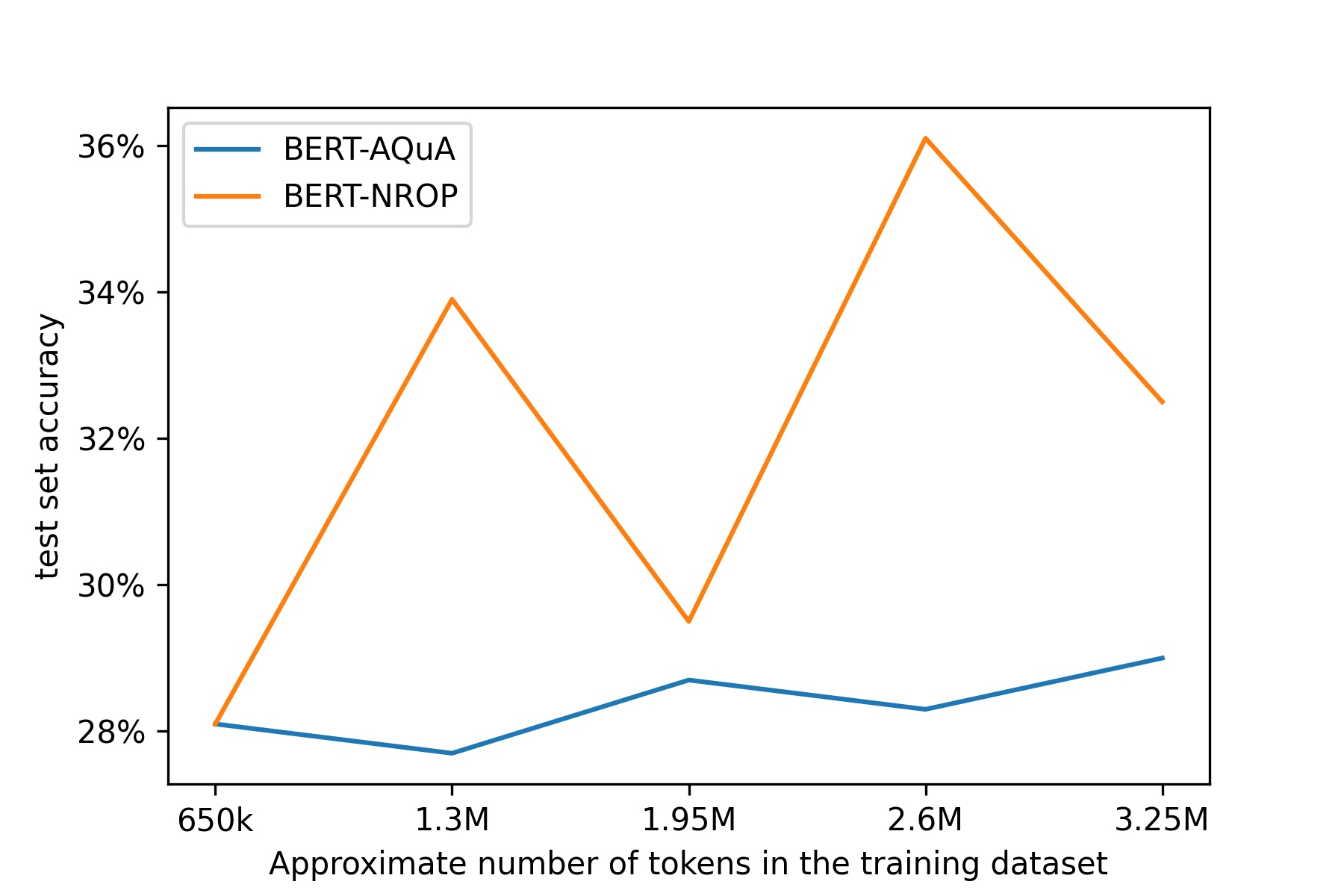} 
    \caption{
    Accuracy scores conditioned on the number of tokens available for training. To support our argument that training on rationales is qualitatively different than questions, we align both together so that we have comparable number of tokens in both cases. Plot shows the progression of the dataset size. Starting with 650K of tokens - 20\% dataset BERT-AQuA and 6.66\% for BERT-NROP and ending with 3.25M - 100\% of dataset for BERT-AQuA and 33.3\% dataset for BERT-NROP. This shows that training with rationales leads to a better representation. Even better than training with more questions. 
    }
    \label{fig:fraction_dataset}
\end{figure}

\begin{figure*}[bt]
\centering
    \includegraphics[width=0.75\linewidth]{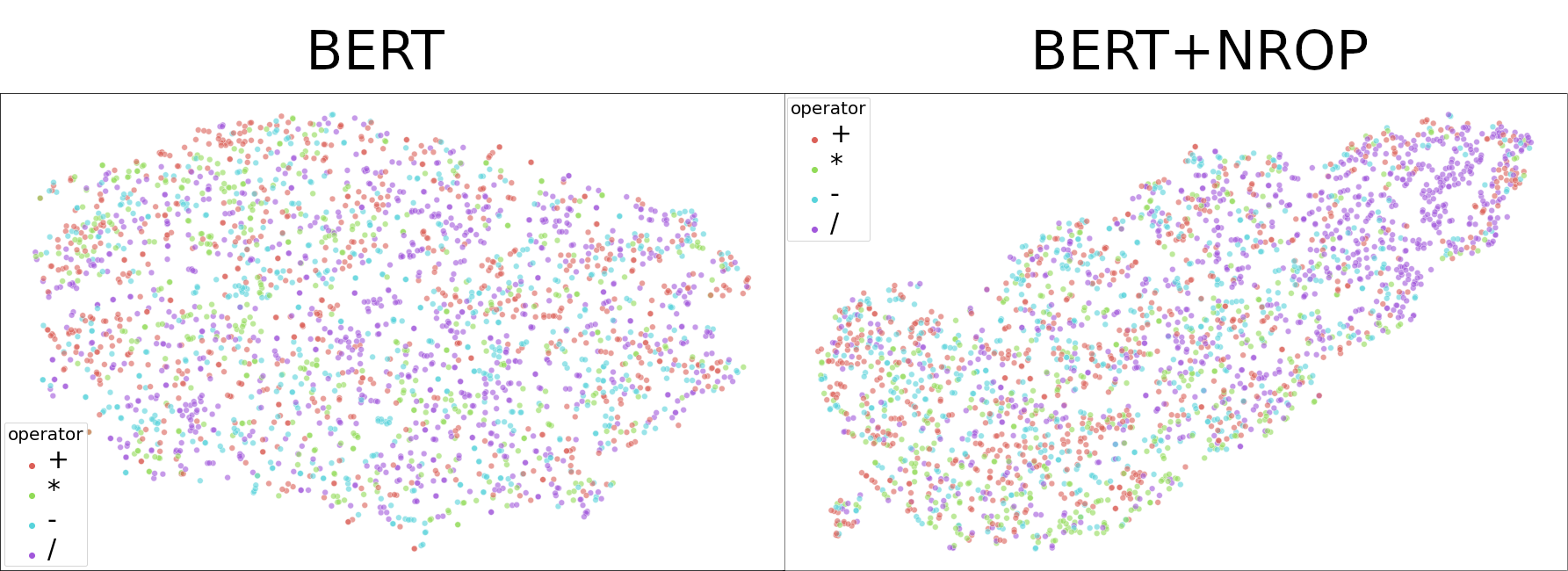} 
    \caption{BERT and BERT+NROP embeddings. Colours represent different operators in rationales (T-SNE). 
    BERT+NROP embeddings better separate operators.
    }
    \label{fig:tsne-both}
\end{figure*}

\noindent
\textbf{Rationales.} 
We hypothesize that rationales contain information that is either missing or hard to extract from questions. For instance, their structure is different; they are more formal with emphasis on the logical steps. However, testing that hypothesis is non-trivial as there is a confounding factor -- adding more rationales results in more data. Therefore, we artificially modify the dataset so that both models (one trained only on questions, and another one on questions and rationales) are trained on roughly the same number of data points. For that, we have estimated that rationales have $1.7$ times more tokens than questions. This means that a question combined with rationale has around $3$ times more tokens than just a question. If our hypothesis is valid, training on $20\%$ questions and rationales should give better results than training on $60\%$ questions (counting the number of tokens). We therefore created samples of respective sizes of just questions and questions combined with rationales. We show our results in Figure~\ref{fig:fraction_dataset}. The results suggest that adding more questions is insufficient and only slightly improves the overall performance. On the other hand, using rationales is more helpful.\newline \noindent
\textbf{Embeddings.} To better understand the difference between BERT and BERT+NROP, we analyze theirs embeddings. For our analysis, we sample 2500 questions with a single operator in rationales, and next we visualise them with T-SNE~\cite{van2008visualizing}. We show both in Figure \ref{fig:tsne-both}. We observe that BERT+NROP embeddings preserve more information about different operators.

\noindent
\textbf{Permutation consistency.} Random guessing on AQuA-RAT yields $20\%$. With that in mind to separate questions that were solved by chance, we have constructed a new evaluation task -- permutation consistency test -- where each question gets 5 answers at different positions. Table~\ref{table:perms} shows our procedure. Here, models only score a single point if they solve all 5 questions correctly. Hence, a random chance is $0.032\%$ in such experiments.

Table~\ref{table:perms_scores} shows our results.
BERT+NROP solves almost three times as many questions as BERT. Additionally, further inspection shows that BERT relies on choosing the answers that most stand out, e.g., numbers ending with zeros or floats while every other option is an integer. We didn't observe that simple patterns with BERT+NROP. Questions solved by BERT+NROP usually contain one or two operations and show that BERT+NROP better understands the problem. 
Below, we exemplify two math problems solved by both models. \newline
\smallskip
{{\footnotesize\it  {\bf Example of a problem solved by BERT+NROP:} 8 man work for 6 days to complete a work. How many men are required to complete same work in 1/2 day?\\
{\bf Answers:} A)93, B)94, C)95, D)96, E)97\\
{\bf Correct Option:} D }
\newline\noindent
{{\footnotesize\it  {\bf Example of a problem solved by BERT} A ship went on a voyage. After it had traveled 180 miles a plane started with 10 times the speed of the ship. Find the distance when they meet from starting point.?\\
{\bf Answers:} A)238, B)289, C)200, D)287, E)187\\
{\bf Correct Option:} C }
\noindent

\begin{table}[bt]
\centering
\resizebox{0.8\linewidth}{!}{
\begin{tabular}{ |p{2.5cm}||p{3.7cm}|  }
 \hline
 \multicolumn{2}{|c|}{Original question} \\
 \hline
\footnotesize{How much is 27 / 3} & \footnotesize{A)13 \textbf{B)9} C)3 D)12 E)17}\\
 \hline
 \multicolumn{2}{|c|}{Generated questions} \\
 \hline
\footnotesize{How much is 27 / 3} & \footnotesize{\textbf{A)9} B)13 C)3 D)12 E)17}\\
\footnotesize{How much is 27 / 3} & \footnotesize{A)13 \textbf{B)9} C)3 D)12 E)17}\\
\footnotesize{How much is 27 / 3} & \footnotesize{A)13 B)3 \textbf{C)9} D)12 E)17}\\
\footnotesize{How much is 27 / 3} & \footnotesize{A)13 B)12 C)3 \textbf{D)9} E)17}\\
\footnotesize{How much is 27 / 3} & \footnotesize{A)13 B)17 C)3 D)12 \textbf{E)9}}\\
 \hline
\end{tabular}

}
\caption{Our generation method for the permutation consistency test. Models get a point only if they solve all them.
}
\label{table:perms}
\end{table}

\begin{table}[bh]
\centering
\resizebox{0.6\linewidth}{!}{
\begin{tabular}{ |p{3cm}|p{1cm}|  }
\hline
\footnotesize{Model} & \footnotesize{Score} \\
 \hline
 \hline
\footnotesize{Random chance} & \footnotesize{0.032\%}\\
 \hline
\footnotesize{BERT} & \footnotesize{4.33\%}\\
 \hline
\footnotesize{BERT+NROP} & \footnotesize{\textbf{11.02\%}}\\
\Xhline{3\arrayrulewidth}
\footnotesize{BERT AUG} & \footnotesize{13.4\%} \\
\hline
\footnotesize{BERT+NROP AUG} & \footnotesize{19.7\%} \\
\hline
\footnotesize{BERT SEP-NC} & \footnotesize{15.0\%} \\
\hline
\footnotesize{BERT+NROP SEP-NC} & \footnotesize{22.7\%} \\
\hline
\footnotesize{BERT SEP-C} & \footnotesize{16.1\%} \\
\hline
\footnotesize{BERT+NROP SEP-C} & \footnotesize{\textbf{23.9}\%} \\
\hline
\end{tabular}
}
\caption{ Our results for the permutation consistency test. }
\label{table:perms_scores}
\end{table}

Drop from $37.0\%$ to $11.02\%$ (Table~\ref{table:perms_scores}) suggests that models rely strongly on the order of answers. To reduce such a bias, we test several permutation invariant losses.

\noindent 1) {\bf AUG}. We sample randomly 25 permutations of all the possible answers and use them during training. Original ordering is not used, so there is no order bias. This is a data augmentation technique.
\newline
\noindent
2) {\bf SEP-NC}. The original models are trained on a 5-class classification task, where we build the representation by using questions and all the candidate answers, i.e., $\textbf{BERT}(Q || P)$. Here, $||$ denotes concatenation, $Q$ is the question and $P$ represents the sequence of all answers. In SEP-NC, we block the path between all the candidate answers and the BERT-base. Next, we use a late-fusion to predict if the given candidate answer matches with the question. That is, we use the following formulation $f(\textbf{BERT}(Q) || \textbf{BERT}(C))$, where $C\in P$ is a single candidate answer and $f$ is a multi-layer perception (with two layers). At test time, the model is prompted to score all five candidate answers and select the one with the highest score. Appendix has more information about this method.
\newline
\noindent
3) {\bf SEP-C}. As models trained with SEP-NC do not have access to all the possible answers, their biases to answer positions are significantly reduced. However, these models cannot compare each answer to all other candidate answers. Here, we use the following formulation $f(\textbf{BERT}(Q||P) || \textbf{BERT}(C))$  to measure the compatibility of the input (question $Q$ and all the candidate answers $P$) with the given candidate answer $C\in P$. We also reset the positional encoding between every possible answer in $P$. In such a way, we hypothesise the network can learn a less biased representation, and on the other hand, use relationship between the candidate answers. Table~\ref{table:perms_scores} shows SEP-NC and SEP-C vastly outperform the original model on the permutation consistency test. Details are in the appendix.

SEP-NC and SEP-C improve permutation consistency tests. Yet, they give similar results to original methods in accuracy measuring task. They achieve respectively $33.5\%$ (SEP-NC) and $35.4\%$ (SEP-C).

\noindent
\textbf{Questions difficulty.} To better understand the models' performance, we check which questions are difficult for the model. We categorize questions by their difficulty for BERT-NROP and BERT. 
To estimate a question's difficulty, we have ranked the candidate answers according to the model's uncertainties.  For instance, if the correct answer has the 2nd largest probability, we assign to that question difficulty two. With that, we group questions into 5 difficulty categories, from the easiest: $D_1, .., D_5$. 

Manual inspection shows that for BERT+NROP: {\color{red} $D_5$} requires additional knowledge or implicitly defined numbers (e.g., adding first 100 numbers), {\color{orange} $D_4$} requires geometry or non-linear equations and systems, {\color{yellow} $D_3$} requires solving linear systems with a few basic operations, {\color{blue} $D_2$} requires solving simple equations, and {\color{green} $D_1$} has one or two basic operations with clearly written numbers. We show an example from each group in the supplementary material. We didn't observe a similar pattern for BERT with the exception of the easiest group {\color{green} $D_1$} where the model chooses the answer that is somewhat different from other candidates.
We provide an example of each group in the supplementary materials.

Finally, we also compare the difficulty of questions with the difficulty perceived by humans. For that, we have conducted a small-group human study, where we have asked participants to solve some AQuA-RAT questions and rate their difficulty. We find a positive correlation between the difficulty measured by our models (as described above) to the difficulty judged by humans. We give more details in the appendix.

\paragraph{Conclusions.} We have investigated if BERT~\cite{devlin-etal-2019-bert} -- a pre-trained, large language model -- can deal with mathematical reasoning. We find that its representation is biased~\cite{brown2020language,bhardwaj2020investigating,kurita2019measuring} also in mathematics. We investigate and describe that bias. Our novel pretext tasks and losses reduce that bias, but the network still finds shortcuts. We hope our work will spark interest of the community in developing language models capable of mathematical reasoning.

{\small
\paragraph{Acknowledgements.} We thank Wang Ling (DeepMind) for his comments and suggestions on our draft. Also, we thank Piotr Biliński and all participants of the 2020 Machine Learning Project course at the University of Warsaw for the conversations about the project. All experiments were performed using the Entropy cluster funded by
NVIDIA, Intel, the Polish National Science Center grant
UMO-2017/26/E/ST6/00622 and  ERC Starting Grant TOTAL. The work of
Henryk Michalewski was supported by the Polish National Science Center
grant UMO-2018/29/B/ST6/02959.
}

\bibliographystyle{acl_natbib}
\newpage 
\appendix

\section*{Impact Statement}
Our research follows the data-driven paradigm for creating general-purpose language models with some mathematical skills. We expect that mathematically aware language models will broaden the spectrum of topics they can understand, increasing their reliability and making them more useful.

Improving mathematical abilities and coherence in language models is likely to affect question-answering or dialogue systems, search engines or text summarization systems. 

One considerable risk in developing language models at scale is that they could use various workarounds and biases to achieve their results. We have shown that issues in the context of mathematical reasoning. Such problems can become hazardous when wrong numbers could lead to bad decisions. Additionally, a person could easily fall into the fallacy that the order of magnitude is correct even if the answer is incorrect. As we showed, the model can favour round numbers over the ones close to the right answer. To mitigate the risk, we encourage considering additional tests and investigating the models more rigorously.

\section{AQuA-RAT example}

\noindent
\smallskip
{{\footnotesize\it  {\bf Question:} A starts a business with Rs.40,000. After 2 months, B joined him with Rs.60,000. C joined them after some more time with Rs.120,000. At the end of the year, out of a total profit of Rs.375,000, C gets Rs.150,000 as his share. How many months after B joined the business, did C join?\\
{\bf Options:} A) 30, B) 32, C) 35, D) 36, E) 40\\
{\bf Rationale:} 
\begin{center}
Assume that C was there in the business for x months \\
$A:B:C = 40000*12 : 60000*10 : 120000*x$ \\ 
$= 40*12 : 60*10 : 120x = 40 : 5*10 : 10x$ \\ 
$=8 : 10 : 2x$ \\
$= 4 : 5 : x$ \\
C's share $= 375000*x/(9+x) = 150000$ \\
$=> 375x/(9+x) = 150$ \\
$=> 15x = 6(9+x)$ \\
$=> 5x = 18 + 2x$ \\
$=> 3x = 18$ \\
$=> x = 18/3 = 6$ \\
It means C was there in the business for 6 months. Given that B joined the business 
after 2 months. Hence C joined after 4 months after B joined \\
Answer is B
\end{center}}
\noindent

\section{Input representation}
 All BERT variants use the representation that corresponds to a special token [CLS] that we put at the beginning of the whole input sequence consisting of question tokens followed by rationale tokens, and in the downstream, question-answering task, rationale tokens are replaced by the answer options. With that, the classification uses the contextual embedding of [CLS] that captures the entire input. MLM classifies over the entire vocabulary of possible words while the other two losses use a binary cross-entropy loss for the predictions. 
\section{Training protocol}
We train all our architectures on AQuA-RAT using the following training phases. In all cases, we choose our best model based on the performance on the validation set (dev set), and report the final performance on the test set.

\paragraph{Pre-training.} Each model is pre-trained on a large corpus of texts written in natural language sampled from English Wikipedia and BooksCorpus~\cite{bert,zhu2015aligning}. We use this as the base (BERT-base) model that is also used in all other variants of BERT. In practice, we initialize all the models with the weights using the HuggingFace library~\cite{Wolf2019HuggingFacesTS} and don't keep final layer for fine-tuning. Our model therefore has the same number of weights as BERT-base. 

\paragraph{Self-supervision.} Here, we use our newly introduced losses, ROP and NROP, where our models use questions and possibly rationales from the AQuA-RAT dataset. Both questions and rationales use the same word embeddings. However, to distinguish between both modalities we use two segment embeddings. The first one for all the question tokens, and the second one for all the rationale tokens. That is, the segment embedding is shared among all the question tokens, and separately among all the rationale tokens.
We use dynamic masking~\cite{liu2019roberta}. Here, tokens are randomly masked for each batch. We naturally extend this approach to other losses that we use in this phase. That is, ROP and NROP negative examples are randomly recreated every $k$ epochs, where $k=2$ in our case. 

\paragraph{Fine-tuning} is the last training phase. Here, once our models have learnt the representation during the self-supervised phase, we tune such a representation to the question-answering downstream task. In this task, our input consists of question tokens and possible answer options. There are five such options that comes with the dataset. Like other methods, we tread this as a five-class classification task where the classification head is added on top of the final embedding of the input. We consider the embedding corresponding to the first (from the left) [CLS] token as such the final representation.

\section{Implementation details}
In our experiments, we use four TITAN V GPUs. We use a multi-gpu setup. In the pre-training phase, we use batch size equals to four for each GPU device. Therefore the effective batch size equals to sixteen. We use the learning rate $5\cdot10^{-5}$ and trained the models for 24 epochs.
In the fine-tuning phase, we use early stopping criteria, based on the accuracy score on the validation set. We use the following criteria. If the model does not improve the performance in 15 consecutive epochs, we stop training, and evaluate a model that yields the highest validation performance. We use ADAM optimizer with learning rate $10^{-5}$ and gradient clipping that sets the maximal gradient's norm to one. 
All our settings use the same hyper-parameters but they differ due to the random initialization of our self-supervised networks (during the self-supervised training phase) and the classification networks (during the fine-tuning phase). Self-supervision phase takes around 4 days on 4 GPUs, whereas fine-tuning takes 8 hours on a single GPU.

\section{Question difficulty}
At this section we present an example from each difficulty group for BERT+NROP and BERT. We have described the grouping procedure in the main paper.
\subsection{BERT+NROP}
\noindent
\smallskip
{{\footnotesize\it  {\bf \color{red}$D_5$:} How many ways A boy can reach the top of stairs which contain 10 steps, when he can take either one or two steps every time? \\
{\bf Answers:} A)88, B)89, C)90, D)91, E)92\\
{\bf Correct Answer:} B \\
{\bf Model Answer:} D}}
\smallskip \newline
\noindent
{{\footnotesize\it  {\bf \color{orange}$D_4$:} A square piece of cloth is trimmed by 4 feet on one edge to form a rectangular piece, which is then cut diagonally in half to create two triangles. If the area of each of triangle is 70 square feet, what was the perimeter (in feet) of the original piece of square cloth?\\
{\bf Options:} A)56, B)58, C)60, D)62, E)64\\
{\bf Correct Answer:} A\\
{\bf Model Answer:} B}}
\smallskip \newline
\noindent
{{\footnotesize\it  {\bf \color{yellow}$D_3$:} Train A leaves a station every 16 minutes and Train B leaves every 17 minutes. If both trains just left the station simultaneously, how long until they do so again? \\
{\bf Options:} A)272 minutes, B)304 minutes, C)190 minutes, D)70 minutes, E)35 minutes\\
{\bf Correct Answer:} A\\
{\bf Model Answer:} B}}
\smallskip \newline
\noindent
{{\footnotesize\it  {\bf \color{blue}$D_2$:} 10kg of a mixture contains 30\% sand and 70\% clay. In order to make the mixture contain equal quantities of clay and sand how much of the mixture is to be removed and replaced with pure sand?  \\
{\bf Options:} A)10/7, B)20/7, C)30/7, D)40/7, E)50/7\\
{\bf Correct Answer:} B\\
{\bf Model Answer:} C}}
\smallskip \newline
\noindent
{{\footnotesize\it  {\bf \color{green}$D_1$:} If one third of 3/4 of a number is 21. Then, find the number? \\
{\bf Options:} A)84, B)66, C)28, D)19, E)11 \\
{\bf Correct Answer:} D\\
{\bf Model Answer:} D}}

\subsection{BERT}
\noindent
\smallskip
{{\footnotesize\it  {\bf \color{red}$D_5$:} The length of the ribbon was originally 30 cm. It was reduced in the ratio 5 : 3. What is its length now? \\
{\bf Answers:} A)18, B)30, C)6, D)15, E)12\\
{\bf Correct Answer:} A\\
{\bf Model Answer:} B}}
\smallskip \newline
\noindent
{{\footnotesize\it  {\bf \color{orange}$D_4$:} An electric pole, 14 metres high, casts a shadow of 10 metres. Find the height of a tree that casts a shadow of 15 metres under similar conditions. 
\\
{\bf Options:} A)21, B)22, C)20, D)23, E)24 \\
{\bf Correct Answer:} A\\
{\bf Model Answer:} C}}
\smallskip \newline
\noindent
{{\footnotesize\it  {\bf \color{yellow}$D_3$:} A rope 20 meters long is cut into two pieces. If the length of one piece of rope is 3 meters shorter than the length of the other, what is the length, in meters, of the longer piece of rope? \\
{\bf Options:}  A)7.5, B)8.9, C)9.9, D)11.5, E)11.7\\
{\bf Correct Answer:} D\\
{\bf Model Answer:} B}}
\smallskip \newline
\noindent
{{\footnotesize\it  {\bf \color{blue}$D_2$:} Jerry purchased a 1-year \$5,000 bond that paid an annual interest rate of 12\% compounded every six months. How much interest had this bond accrued at maturity?  \\
{\bf Options:} A)\$5102, B)\$618, C)\$216, D)\$202, E)\$200\\
{\bf Correct Answer:} B\\
{\bf Model Answer:} A}}
\smallskip \newline
\noindent
{{\footnotesize\it  {\bf \color{green}$D_1$:} I have a money pouch containing Rs. 700. There are equal number of 25 paise coins, 50 paise coins and one rupee coins.
How many of each are there?  \\
{\bf Options:} A)453, B)651, C)400, D)487, E)286 \\
{\bf Correct Answer:} C\\
{\bf Model Answer:} C}}

\section{Permutation invariant methods}
In the main paper, we have shown that typical models can use positional biases in achieving answers. This results in a low permutation consistency score (Table 3 in the main paper). To handle that issue, we have defined extra variants that do not use positional encodings for the answer options and instead they rely on the retrieval mechanics where input representations are matched against the candidate answers. Here, we describe two such variants.

\subsection{Original methods}
Original models create an embedding of a sentence extended by possible questions. This embedding is then transformed by a linear layer to predict the correct answer. That is, 
\[ o_1 = f_1(\textbf{BERT}(Q || P)) \]
where $o_1$ is a 5-dimensional vector with probabilities for each possible answer, $Q$ is a question, $P$ are all possible answers, $ || $ represents concatenation, $f_1$ is a single fully connected layer from 768-dimensional space to 5-dimensional space with the softmax activation. $\textbf{BERT}$ is a BERT-base sentence embedding.  The same approach is used for BERT+(N)ROP.

\subsection{SEP-NC}

In SEP-NC and SEP-C, we use separate embeddings for a question and \textbf{SEP}arate embedding for a candidate answer. They differ, however, in the fact that SEP-C has access to all five possible answers, while SEP-NC has access only to one prompted candidate answer. Therefore NC stands for "no candidates", while C stands for "candidates".

We train the SEP-NC model on a binary classification task to predict whether each candidate answer $C$ is correct. The method produces two embeddings, one for question and another one for a candidate answer $C\in P$, and next concatenates them. That is,
\[o_2 = f_2(\textbf{BERT}(Q) || \textbf{BERT}(C))\]
where $o_2$ is an estimated probability that $C$ is a correct answer, $P$ is the sequence of all possible answers, $f_2$ is a single fully connected layer from 1536 (768 * 2) dimensional space to 1-dimensional space with the sigmoid activation. Note that, all candidate answers are independent of the question. That is, BERT cannot use positional biases in deriving an answer. At test time, the model is prompted to score all five candidate answers and select the one with the highest score.  We naturally extended that approach to BERT+ROP and BERT+NROP. Table 3 (the main paper) shows a significant improvement over the baseline method.

\subsection{SEP-C}
SEP-NC method could be too restrictive as it does not allow the model to compare against different answers. Therefore, we propose another approach that 1) alleviate the issue with positional biases, but 2) can compare between different answer options. We call that approach SEP-C.

Originally for each token, a positional encoding is assigned based on its position.
In SEP-C, before assigning positional encoding, we artificially reset the position at the beginning of each possible answer. For example, if possible answers are: $a) 10, b) 20, c) 30, d) 40, e) 50$ they are changed into $10;20;30;40;50$ and after the tokenization, we get the following list of tokens: ['1','0', ';', '2', '0', ';', '3', '0', ';' ,'4', '0', ';', '5', '0']. Modified positional encoding will assign value based only on the relative position to the beginning of the current possible answer. Therefore, in the example above, each '0' will receive the same positional encoding, and '1' will get the same positional encoding as '2', '3', and so on. 

Formally, we have
\[o_3 = f_3(\textbf{BERT}(Q||P_m) || \textbf{BERT}(C))\]
where $P_m$ is the sequence of all the possible answers but modified as explained above. Note that, in this formulation, the model can use the information for all the possible answer options, but their order is not taken into account.
Table 3 (the main paper) shows a significant improvement over the baseline method.

\subsection{Human study}
We carried an initial human study on the group of 16 volunteers from University of Warsaw. Volunteers were Mathematics and Informatics students from the Faculty of Mathematics, Informatics and Mechanics. We asked the participants to solve questions sampled from the AQuA-RAT dataset. We are interested in the relation between BERTs difficulty, BERT+NROP difficulty and human difficulty. Therefore to have a full image we would like to have 2 questions for each question difficulty pair, for example ({\bf \color{green}$D_1$} BERT, {\bf \color{blue}$D_2$:} BERT+NROP). However, that would give 25 combinations and 50 questions if we wanted to have 2 questions per combination. That would be too much to ask from a volunteer participant. In order to reduce the number of questions, we group our 5 difficulty groups into 3 categories as follows. 

\begin{itemize}
    \item Easy:  {\bf \color{green}$D_1$} 
    \item Medium: {\bf \color{blue}$D_2$} and {\bf \color{yellow}$D_3$} combined
    \item Hard: {\bf \color{orange}$D_4$} and {\bf \color{red}$D_5$} combined
\end{itemize}

Because of that we have only 9 possible combinations and by sampling 2 questions from each combination we still have a feasible number of questions (18).

Apart from solving the question, we asked to rate question difficulty on a scale from 1 (the simplest) to 10 (the most challenging). In general, our participants were knowledgeable in math and solved all the questions correctly. 
With that grouping we now 

The average human-rated difficulty for each of 9 combinations is presented in ~\autoref{fig:human}. The results show that the progression of human difficulty is correlated with the difficulty judged by the models. Additionally, the human difficulty seems to be more sensitive to BERT+NROP difficulty than to BERTs. In other words, increasing the difficulty of BERT+NROP will increase the human difficulty more than the increasing difficulty of BERT. This observation fits our previous observations that BERT+NROP solves the most straightforward questions while BERT is looking for some leaks, like looking for the roundest answer.

\begin{figure}[t!]
\centering
    \includegraphics[width=1\linewidth]{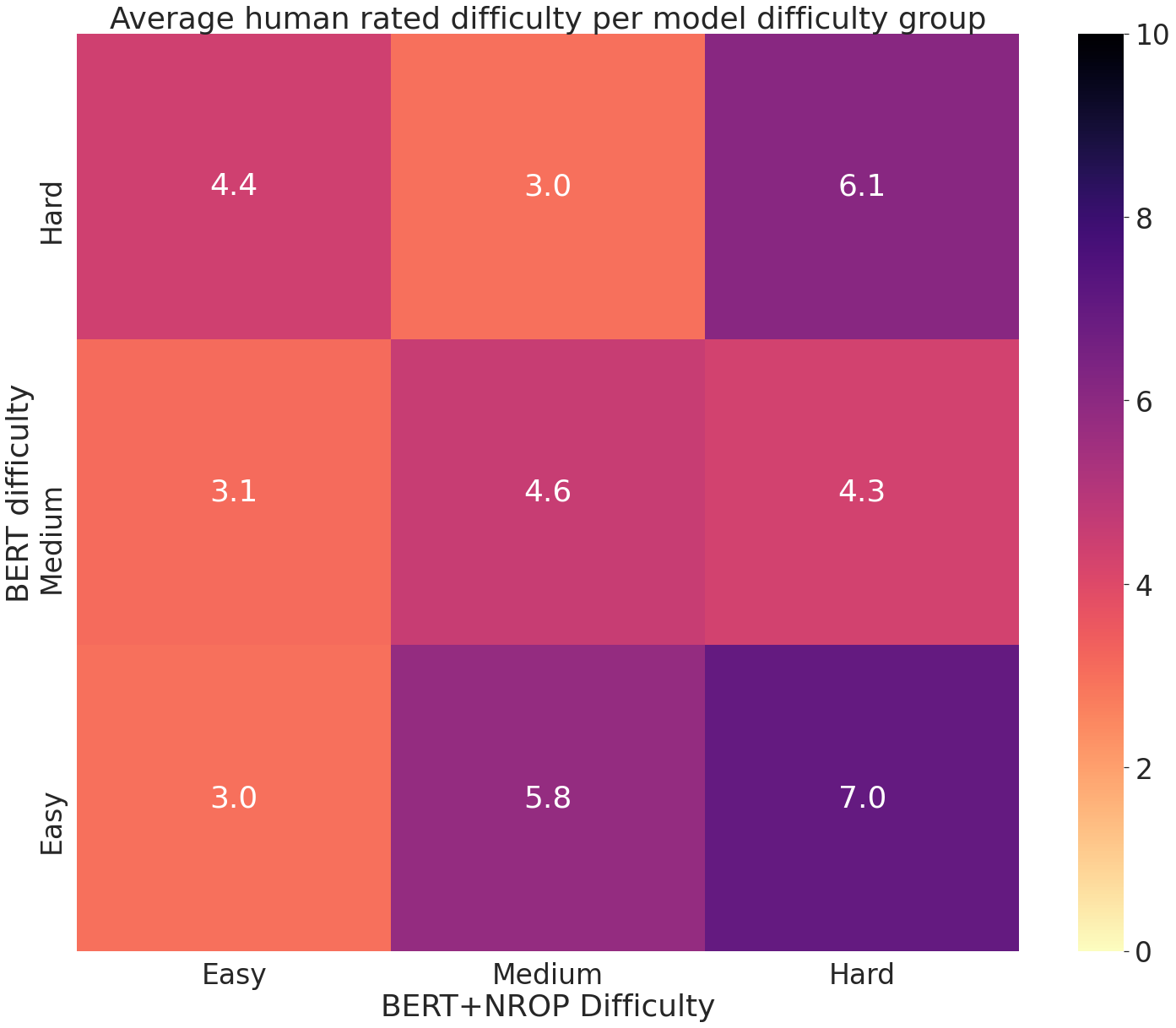} 
    \caption{The average human-judged difficulty for questions from each model difficulty group.
    }
    \label{fig:human}
\end{figure}

\section{Distribution of answers}
Table~\ref{table:answers_distribution} shows the distribution of the answers in the AQuA-RAT~\cite{aqua} dataset in all the folds. Imbalance in distributions could potentially be used by models to find easy, shortcut solutions. For instance, a constant classifier that always choose the first answer (A) gets about $24\%$ test accuracy.

\begin{table}[bt]
\centering
\resizebox{1\linewidth}{!}{
\begin{tabular}{ |p{2cm}||p{1.5cm}|p{1.5cm}| p{1.5cm}| p{1.5cm}| p{1.5cm}|   }
 \hline
\footnotesize{dataset} & \footnotesize{A} & \footnotesize{B} & \footnotesize{C} & \footnotesize{D} & \footnotesize{E}\\
 \hline
\footnotesize{train} & \footnotesize{21.03\%} & \footnotesize{22\%} & \footnotesize{22.87\%} & \footnotesize{19.95\%} & \footnotesize{14.15\% }\\
 \hline
\footnotesize{dev} & \footnotesize{27.17\%} & \footnotesize{25.98\%} & \footnotesize{16.93\%} & \footnotesize{19.69\%} & \footnotesize{10.24\$ }\\
 \hline
\footnotesize{test} & \footnotesize{24.80\%} & \footnotesize{22.83\%} & \footnotesize{20.87\%} & \footnotesize{18.11\%} & \footnotesize{13.38\% }\\
 \hline
\end{tabular}
}
\caption{Answer distribution in each dataset.
}
\label{table:answers_distribution}
\end{table}

\section{Negative results}
While developing our self-supervised losses, we have developed another loss that turned out to be unhelpful. Here, we describe that loss as some its parts could be insightful for others. (N)ROP is a local loss focusing on rationales but not on the connections between questions and rationales. For that, we have developed Question Rationale Alignment (QRA). QRA changes a rationale with $50\%$ probability to a randomly chosen rationale from the current batch. However, simply changing rationales would result in trivially solvable task in most cases. All the model would have to do is check whether numbers in the rationale and the question match. Hence, we mask number tokens with a special token
QRA alone or QRA combined with NROP does not improve the results, it gives it gives 33.9\% accuracy on the test in the best combination, so we didn't include it in the main results.

\section{Related work}
We are inspired by the following research.
\newline\noindent
\textbf{BERTology.}
We use BERT~\cite{devlin-etal-2019-bert} as our core. It uses Transformers~\cite{vaswani2017attention}; powerful neural architectures that applies a trainable function to all the pairs of input embeddings. It also uses masking that covers a fraction of the input words and requires the network to predict the hidden words based on the context. 
With both ingredients, the meaning (representation) of a word emerges from the ``company it keeps''~\cite{firth1961papers}. In practice, often, such representations are pre-trained on large textual corpora with no need for annotations, and next fine-tuned on the downstream tasks. BERT's strong performance has resulted in the Cambrian explosion of studies of the inner working mechanisms and various modifications~\cite{clark-etal-2019-bert,de2019bertje,lan2019albert,liu2019roberta,sanh2019distilbert,radford2018improving,raffel2019exploring,yang2019xlnet}. Finally, our Reasoning Order Prediction (ROP) is inspired by Sentence Order Prediction (SOP)~\cite{lan2019albert}. However,  ROP works with multiple rationale sentences, where by changing the order we force the network to understand the consecutive ``reasoning'' steps. We have also further extended ROP to a more difficult Neighbor Reasoning Order Prediction (NROP).
\newline\noindent
\textbf{Language and math.}
Development psychologists~\cite{cocking1988linguistic,mestre2013role} often argue for the necessity of learning languages and point out that those with limited language skills are in danger of under-performing at school. Moreover, it is also believed that language studies involve discipline in learning and manipulating formal structures, and thus may promote the development of the organization of thoughts also required in mathematical reasoning. The similarity between linguistic competence and mathematics is especially pronounced when solving math word problems~\cite{fuchs2006cognitive,fuchs2008problem,wang2016cognitive}. Interestingly, attention appears to be crucial in problem solving~\cite{fuchs2006cognitive,pasolunghi1999working}.~\cite{crossley2017linking} show that language skills are correlated with the performance in mathematical tests also among the university students. In particular, they pointed out that ability to use complex syntactic structures and cohesion devices are linked to better scores in a blended discrete mathematics course. We take inspiration from all such studies and decide to build our mathematical model based on language models.
\newline\noindent
\textbf{Math word problems.}
Solving math word problems is a significant component of the mathematics curriculum and is taught very early, thoroughly, and universally. Such the emphasize is often motivated by that solving them is among the best predictors of employability, and is considered as a distinct area of mathematical competence~\cite{murnane2001different,wang2016cognitive}. Since solving such problems is unique to human intelligence, math word problems are also interesting for the AI community. This results in various approaches, more traditional symbolic methods, neural networks, and neuro-symbolic methods.
~\cite{bobrow,charniak1969computer,shi2015automatically,aqua,amini-etal-2019-mathqa,parisotto2016neuro,wang2018mathdqn,zou-lu-2019-text2math} as well as datasets~\cite{aqua,amini-etal-2019-mathqa,huang-etal-2016-well,saxton2019analysing}
An interesting approach is proposed in ~\cite{rabe2020mathematical}, in which authors use self-supervised tasks on parsing trees of formal expressions. This approach requires syntax trees, and hence we would have to use an external parser. As our goal was to make an end to end model, we did not experiment with it, but there are no obstacles against using it in symbiosis with our methods.
~\cite{geva-etal-2020-injecting} also proposes self-supervised training for improving mathematical abilities in language models. We, however, focused on a data-driven approach to exclude choice biases and therefore restricted ourselves from using generated data.
\newline\noindent
\textbf{Rationales.}
In human communication, we always expect there is some rationale behind each decision. Hence, we set the same expectations to our artificial agents. Symbolic or semi-symbolic architectures naturally produce justifications as a sequence of formulas in some formal language~\cite{lane2005explainable,core2006building,lomas2012explaining,johnson1994agents,liang2016learning,malinowski2014multi}.  Ideally, such rationales would also be shared and communicated to us through some language. The latter approach is especially appealing when applied to black-box neural networks. For instance,~\cite{hendricks2016generating} propose a system that classifies the input image as well as it produces a textual explanation on ``why this class is suitable for the given image''. \\
\indent
Systems that produce explanations either in the form of the language~\cite{aqua,hendricks2016generating}, attention~\cite{bahdanau2014neural,mnih2014recurrent,gulcehre2016dynamic,malinowski2018learning,xu2016ask,yang2016stacked}, phrase selection~\cite{lei2016rationalizing}, distillation into programs~\cite{hajipour2020ireen}, or decision trees~\cite{alaniz2019explainable} can potentially increase the transparency of the black-box neural networks.
However, most of these approaches create rationales posthoc where the justification is conditioned on answers or by querying the network. In our work, we use rationales to learn a finer representation that can potentially lead to better decisions. In this sense, our technique is conceptually closer to methods that derive answers based on the program and use rationales paired with questions to guide the program induction process~\cite{aqua}.

\newpage
\bibliography{references, anthology}

\end{document}